\documentclass[default,iicol]{sn-jnl}

\usepackage{graphicx}
\usepackage{caption}
\usepackage{subcaption}
\jyear{2021}%

\theoremstyle{thmstyleone}%
\theoremstyle{thmstyletwo}%

\theoremstyle{thmstylethree}%

\begin{document}

\title[Article Title]{How I learned to stop worrying and love the curse of dimensionality: an appraisal of cluster validation in high-dimensional spaces}


\author{\fnm{Brian} \sur{Powell}}\email{brian.a.powell@jhuapl.edu}

\affil{\orgname{The Johns Hopkins University Applied Physics Laboratory}, \orgaddress{\street{11100 Johns Hopkins Rd.}, \city{Laurel}, \postcode{20723}, \state{MD}, \country{USA}}}


\abstract{The failure of the Euclidean norm to reliably distinguish between nearby and distant points in high dimensional space is well-known. This phenomenon of distance concentration manifests in a variety of data distributions, with iid or correlated features, including centrally-distributed and clustered data.  Unsupervised learning based on Euclidean nearest-neighbors and more general proximity-oriented data mining tasks like clustering, might therefore be adversely affected by distance concentration for high-dimensional applications.  While considerable work has been done developing clustering algorithms with reliable high-dimensional performance, the problem of cluster validation---of determining the natural number of clusters in a dataset---has not been carefully examined in high-dimensional problems.  In this work we investigate how the sensitivities of common Euclidean norm-based cluster validity indices scale with dimension for a variety of synthetic data schemes, including well-separated and noisy clusters, and find that the overwhelming majority of indices have improved or stable sensitivity in high dimensions.  The curse of dimensionality is therefore dispelled for this class of fairly generic data schemes.}




\maketitle

\section{Introduction}\label{sec1}
The notion of proximity is essential for a variety of unsupervised learning techniques, including nearest-neighbors-based analyses and clustering problems.  It is known, however, that Euclidean distances tend to ``concentrate'' in high dimensions \cite{Beyer,Hinneburg,Pestov} with the result that data points tend towards equidistance.  This phenomena threatens the ability of algorithms based on Euclidean distance to perform when feature spaces become large.  Along with the exponential increase in time complexity of high dimensional problems, distance concentration has earned its place as a pillar of the ``curse of dimensionality''.  

Quantitatively, distance concentration is exhibited as the limit
\begin{equation}
\lim_{d \rightarrow \infty} \frac{D_{\rm max}}{D_{\rm min}} = 1,
\end{equation}  
where the distances to the closest ($D_{\rm min}$) and farthest ($D_{\rm max}$) points from a particular reference point are driven to equality as the dimensionality, $d$, of the feature space increases.  While the extent to which this phenomenon is realized depends on the nature of the data distribution, it is known to occur for independent and identically distributed (iid) data distributions with finite moments, as well as data with correlated features \cite{Beyer}; this includes both uniformly-distributed and clustered data \cite{Hinneburg}.  
While there exist data distributions that do not exhibit distance concentration, like linear latent variable models \cite{Durrant}, the curse of dimensionality has been demonstrated to affect a varied collection of both real-world and synthetic datasets \cite{Flexer2015,Kumari,Kaban} (see \cite{Hall,Ahn} for the required conditions for a dataset to exhibit distance concentration and related geometrical phenomena.)  

\begin{figure}[ht]
\centering
\includegraphics[width=3in]{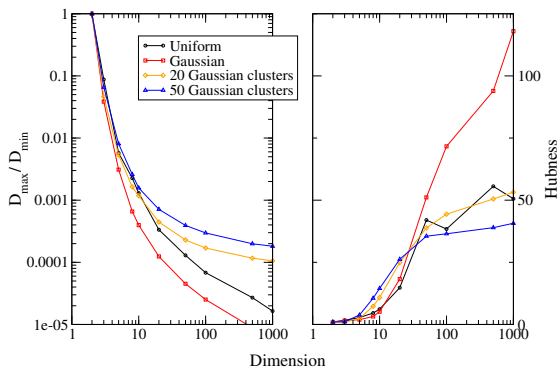}
\caption{\footnotesize{Behavior of $D_{\rm max}/D_{\rm min}$ (left) and hubness (right) as a function of dimension for four types of data distributions: uniform and Gaussian, and datasets with 20 and 50 Gaussian clusters.}}
\label{dist}
\end{figure}
A related distance-oriented phenomenon in high-dimensional spaces is the formation of {\it hubs}: points that are close to a large fraction of the dataset \cite{Radovanovic09,Radovanovic10}.  The presence of hubs can result in highly unbalanced clustering, with a small number of clusters containing a disproportionate amount of data \cite{Tomasev14,Schnitzer15}, and can render nearest neighbors-based analysis meaningless;  like distance concentration, hubness is apparently a fairly generic occurrence in high-dimensional datasets \cite{Flexer2015}.  To compute hubness, one counts the number of times each point appears in the $n$-neighborhood of all other points.  The skewness of this distribution, parameterized by $n$, is defined as the hubness.  

To illustrate these phenomena, we plot the ratio $D_{\rm max}/D_{\rm min}$ and hubness with $n=5$ as a function of dimension for a variety of distributions in Figure \ref{dist}.  
Gaussian-distributed data exhibits both phenomena more strongly than uniformly-distributed data, with clustered data least susceptible.
 
Many avenues have been explored to combat and ameliorate these effects, including the use of similarity measures based on the notion of shared nearest neighbors \cite{Yin,Houle10} to perform clustering in high dimensional spaces. 
A substantial amount of work has been put towards the investigation of alternative distance metrics, like the L1 norm and fractional Minkowski distances, with allegedly improved high-dimensional discrimination \cite{Aggarwal2001,Aggarwal22001,Doherty2004,Verleysen05,Hsu,Jayaram}, although the recent work of \cite{Mirkes} finds no such benefit.  Methods to remove hubs via distance scaling have been explored \cite{Schnitzer12,Schnitzer14} as have the use of hub-reducing distance measures \cite{Suzuki};  shared nearest-neighbors have also been shown to reduce hubness \cite{Flexer}.  Dimensionality reduction, a popular set of techniques for doing away with irrelevant and confounding features, has been used to simply avoid the problem all together \cite{Espadoto,Jin,Berchtold}. 

The present work explores the effect of distance concentration on clustering in high-dimensional spaces.  There has been substantial research into the development of ``dimensional-tolerant'' algorithms \cite{Steinbach,Bouveyron,Zimek1,Assent,Pandove,Echihabi}, including subspace and projective methods \cite{Aggarwal2000,Fern,Parsons,Kriegel09,Moise,Kelkar} and modifications to classic clustering algorithms \cite{Hornik,Wu14,Chakraborty}.  These methods are intended to outperform conventional algorithms like $k$-means (and other Euclidean distance-based methods) when the number of features becomes large.  However, very little attention seems to have been paid to the problem of cluster validation---the determination of the number of natural groupings present in a dataset---in high-dimensional spaces. The optimal value of various {\it relative validity criteria}, like silhouette scores, are commonly used to select the most appropriate number of partitions in a dataset; however, these indices are overwhelmingly based on Euclidean distance.  If we are suspicious of the performance of Euclidean-based clustering algorithms, we should be equally concerned with the use of similarly implicated validation measures.  This paper is a study of the performance of a wide variety of such validity criteria as the dimensionality of the feature space grows. 

The most recent, and apparently sole, investigation of this question is the 2017 work of Tomasev and Radovanovic \cite{Tomasev}.  That work examined how 18 cluster quality indices scale with increasing dimension.  The author's reason as follows: if a particular index is maximized at the optimal partition (with $K=k^*$ clusters), for example, then the growth of this index with dimension indicates that it ``prefers'' higher dimensional distributions, and conversely for indices that are minimized at the optimal partition.  

While this is true on its face, it does not tell us whether a particular index is still {\it useful} in high dimensions.  To be of use, the practitioner must be able to identify the optimum of the index over some range of partitionings of the data, and this depends not only on the magnitude of the index at $K=k^*$, but also on its values in the neighborhood of the optimum (at $K=k^*+1$ and $K=k^*-1$).  As an example, suppose some index, $m$, is maximized at $K=k^*$ and that this value grows with dimension: unless the values at $K=k^*+1$ and $K=k^*-1$ scale identically (or less strongly), the curvature of the function $m(K)$ around $K=k^*$ will lessen and the optimum will be less prominent.  This is quite possible in practice, since the scaling behavior of $m(k^*)$ and $m(k^*\pm 1)$ are in general different.  We refer to this ``sharpness'' of the optimum as the {\it sensitivity} of the index.

In this work we assess sensitivity as a function of dimension for 24 relative validity criteria from the standard literature for a variety of different simulated data schemes, including well-separated clusters as well as distributions with noise. We find that all but one of the indices either have improving or stable sensitivity as dimensionality increases for well-separated, Gaussian clusters. This observation agrees broadly with the work of \cite{Tomasev}, but we find that several of the indices that were stable in that analysis (C-index, Cali\'{n}ski-Harabasz, and RS)  actually degrade (C-index) or improve (Cali\'{n}ski-Harabasz and RS).  A good number of indices also perform well on noisy data in high dimensions, and we observe that fractional Minkowski metrics might offer some improvements in some cases. In summary, this analysis concludes that relative validity criteria are {\it unaffected by distance concentration in high-dimensional spaces}.

Before continuing, a note on terminology. In the following, we refer to a {\it cluster} as one of the partitions found by a clustering algorithm.  The number of clusters is a free parameter in most algorithms (the $k$ in $k$-means).  We refer to a {\it grouping} as the set of points arising from a single Gaussian distribution.  The number of groupings is property of the dataset, and so optimal number of clusters is therefore equal to the number of groupings.   

\section{Index sensitivity}
\begin{figure*}[ht]
\centering
\includegraphics[width=6in]{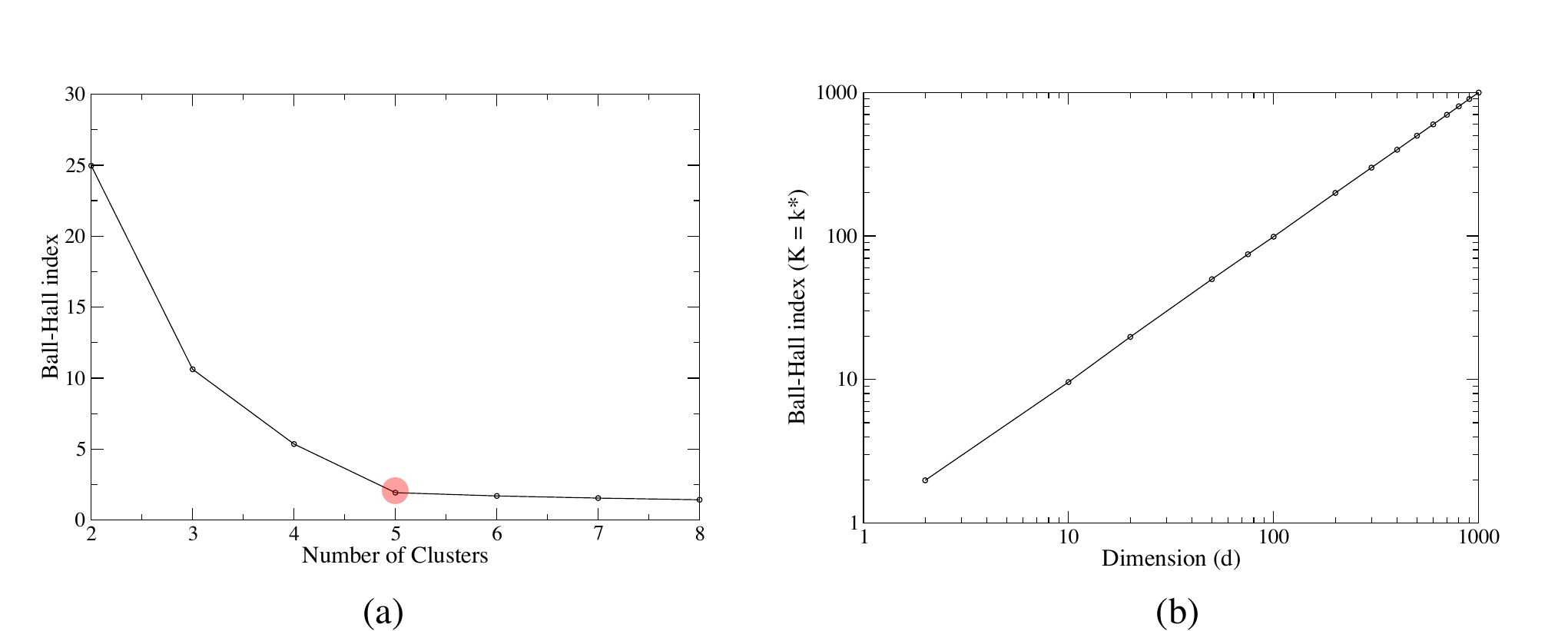}
\caption{\footnotesize{(a) Ball-Hall index as a function of number of clusters, $K$, for a synthetic dataset with $k^* = 5$ well-separated Gaussian clusters in $d=2$ dimensions. The correct number of clusters occurs at the ``elbow'' of this plot. (b) Ball-Hall index evaluated at $K = k^* = 5$ as a function of dimension, $d$, for well-separated Gaussian clusters, showing linear scaling behavior.}}
\label{BH}
\end{figure*}
The sensitivity of an index, $m$, is a measure of the prominence of the optimum of the function $m(K)$.  We introduce this notion of sensitivity within the context of a few examples.

A fundamental quantity in clustering analysis is the {\it within-cluster dispersion}, which is the sum of the squared distances of all points, $x_i$, from the cluster barycenter, $\overline{x}_k$.  It is often written WSS, for {\it within-cluster sum of squares},
\begin{equation}
{\rm WSS}_k = \sum_{x_i \in C_k} \lvert\lvert x_i - \overline{x}_k \rvert\rvert^2.
\label{wss}
\end{equation}
${\rm WSS}_k$ measures how tightly a cluster's points are distributed about its center, and so furnishes a simple measure of relative cluster validity: if a given partition uses too few clusters, then some of these clusters will span multiple distinct groupings and will thus have larger mean WSS values than those clusters spanning single groupings.  Alternatively, if a partition uses too many clusters, then all cluster points will be relatively close to their respective barycenters, and the mean WSS values will be approximately equal across the clusters.  The mean dispersion, averaged over all clusters, plotted as a function of the number of clusters, $K$, thus resembles an ``elbow'' which can be used to determine the optimal number of clusters admitted by a dataset, Figure \ref{BH} (a).  

The {\it Ball-Hall index} is precisely this quantity \cite{Ball},  
\begin{equation}
{\rm BH}_K = \frac{1}{K}\sum_{k=1}^K \overline{{\rm WSS}_k}.
\end{equation}
We would like to understand how indices like Ball-Hall perform as the dimensionality, $d$, of the dataset is increased.  If we assume that our data is arranged into $k^*$ well-separated, Gaussian groups, we can actually make some headway analytically.  Suppose we partition the data into $K = k^*$ clusters.  Analyzing each cluster separately, we can always shift its barycenter to zero; then, each term $\lvert \lvert x_i - \overline{x}_k \rvert\rvert^2 = \lvert \lvert x_i \rvert\rvert^2$ in Eq. (\ref{wss}) is the sum of $d$ squared, mean-zero Gaussians and so is $\chi^2$-distributed with $d$ degrees of freedom.  The mean dispersion, $\overline{{\rm WSS}_k}$, is then the mean of $n_k$ iid $\chi^2$ random variables, which is equal to $d$.  The Ball-Hall index, which is the mean of these means, therefore scales linearly with dimension, Figure \ref{BH} (b).  Since smaller WSS values indicate higher-quality clusters, the growth of WSS with dimension might suggest that the WSS (and indices based on it, like Ball-Hall) degrade as the dimensionality of the dataset increases.  These are the kinds of scaling arguments made to assess the high-dimensional performance of various relative indices in \cite{Tomasev}.  But, this is not the whole story. 

In practice, relative validity indices yield the preferred number of clusters, $K=k^*$, either as an optimum or as a knee- or elbow-point\footnote{These latter types are referred to as {\it difference-like} in \cite{Vendramin} because they can be converted to an optimization problem by taking suitable differences of consecutive values.} when the index is plotted as a function of $K$.  Good indices develop strong optima at $K = k^*$ so that the correct number of clusters can be confidently and reliably determined.  We refer to this property as the {\it sensitivity} of the index to the correct number of clusters, defined simply as the height (or depth) of the optimum relative to its neighboring points: 
\begin{equation}
s = \lvert m_{K=k^*} - {\rm opt}(m_{K=k^*-1},m_{K=k^*+1})\rvert,
\label{sens}
\end{equation}
for generic index $m$ and where opt refers to the maximum or minimum, as the case may be. It is this sensitivity that we would like to study as a function of increasing dimensionality.  

\begin{figure}[ht]
\centering
\includegraphics[width=3in]{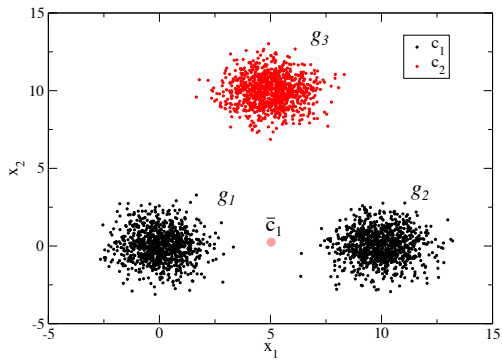}
\caption{\footnotesize{Example partitioning of three natural groupings into two clusters using $k$-means. The barycenter of cluster $c_1$ lies at the midpoint of the two groupings, $g_1$ and $g_2$.}}
\label{cluster_ex}
\end{figure}

\begin{figure*}[ht]
\centering
\includegraphics[width=6in]{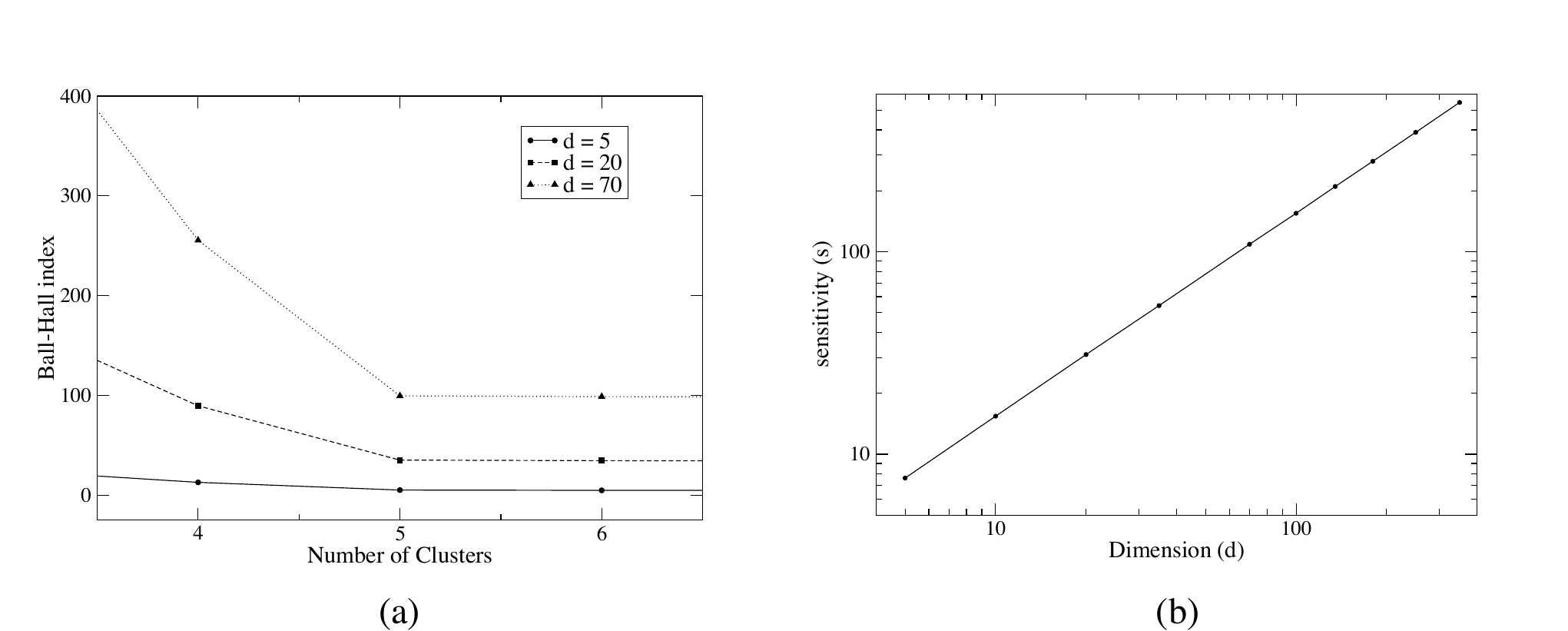}
\caption{\footnotesize{(a) Ball-Hall index as a function of $K$ for data of three different dimensionalities, $d = 5$, 20, and 70, illustrating the increase in elbow steepness with dimension.  (b) The sensitivity of the Ball-Hall index, which is proportional to this elbow steepness, as a function of dimension, showing a linear scaling behavior.}}
\label{BH_steps}
\end{figure*}

Returning to our example of the Ball-Hall index,  as an elbow-like index, ${\rm BH}_{K=k^*}$ is most different from ${\rm BH}_{K=k^*-1}$, since ${\rm BH}_{K=k^*} \approx {\rm BH}_{K=k^*+j}$ for $j > 0$ (these points lie along the horizontal ``forearm''), and so $s = \lvert {\rm BH}_{K=k^*} - {\rm BH}_{K=k^*-1}\rvert$.  We just demonstrated that ${\rm BH}_{K=k^*}$ scales linearly with dimension, and so we must now determine how $ {\rm BH}_{K=k^*-1}$ scales. For ease of exposition, we consider the case of $k^* = 3$, and study ${\rm BH}_{K = 2}$.  An example of how $k$-means with two clusters tends to partition the data is shown in Figure \ref{cluster_ex}.  Notice that two natural groupings, $g_1$ and $g_2$, are spanned by a single cluster, $c_1$: if we shift the cluster barycenter to zero, the sum Eq. (\ref{wss}) for this cluster contains squared Gaussians from two different distributions with means $\mu_1$ and $\mu_2$, each nonzero. The sum of squares of points from $g_1$ follow a {\it noncentral} $\chi^2$ distribution with mean $d(1+\mu_1^2)$, and similarly for points from $g_2$. If we suppose that $\mu_1 = \mu_2 =\mu$ and if we assume groups of equal size, $\overline{{\rm WSS}_{c_1}} \propto d(1+\mu^2)$.  
Meanwhile, $\overline{{\rm WSS}_{c_2}} \propto d$ since $g_3$ is a zero-mean Gaussian and so distances are $\chi^2$-distributed with mean $d$, as per our earlier discussion.  The BH index is the average of clusters $k=1$ and $k=2$, ${\rm BH}_{K=2} \propto d(1+\mu^2/2)$.  With ${\rm BH}_{K = 3}$ for $k^* =3$ scaling linearly with dimension, the sensitivity of the Ball-Hall index is $s_{\rm BH} = \lvert{\rm BH}_{K = 3} - {\rm BH}_{K = 2} \rvert \propto d \mu^2 /2$.  As dimensionality increases, the sensitivity of the Ball-Hall index {\it likewise increases} at a linear rate (cf. Figure \ref{BH_steps}), contrary to any supposition based on the scaling behavior of ${\rm BH}_{K = k^*}$ alone.
While Euclidean distances concentrate even for well-defined clusters in high-dimensions, the sensitivity of the Ball-Hall index is unaffected: sensitivity, as the difference of average $\overline{\rm WSS}_k$ applied to two different partitions, each unbounded and scaling linearly with $d$, also grows linearly in $d$ without bound.   This result will hold true for any index defined in proportion to WSS in this way.  

The quantity ${\rm WSS}_k$ can be thought of as describing the intra-cluster structure of the dataset.  In this sense, it probes a single scale: those distances of order the cluster diameter.  Another fundamental quantity, the {\it between-cluster dispersion}, or BSS (for between-cluster sum of squares), measures the degree of separation of different clusters,
\begin{equation}
{\rm BSS} = \sum_{k=1}^K n_k \lvert \lvert \overline{x}_k - \overline{x} \rvert \rvert ^2,
\label{bss}
\end{equation}
where $n_k$ is the number of points in cluster $k$ and $\overline{x}$ is the barycenter of the entire dataset.  Importantly, BSS probes a {\it different scale} than WSS, and so provides new and separate information about the data partition.  Many of the more robust relative validity criteria incorporate two or more quantities like these that operate on different scales.  Whereas indices based on ${\rm WSS}_k$ (like BH) essentially inherit its scaling behavior, once a second quantity is introduced there is great variety in how the resulting index scales, depending on how the base quantities are combined.  Consider the index of Hartigan \cite{Hartigan},
\begin{equation}
{\rm H} = \log\left(\frac{{\rm BSS}}{{\rm WSS}}\right),
\end{equation}
based on the ratio of inter- and intra-cluster dispersions.  Here ${\rm WSS} = \sum_k {\rm WSS}_k$ is the {\it pooled} within-cluster sum of squares. The optimal number of clusters, $k^*$, corresponds to the knee-point of ${\rm H}_K$ with $H_{K<k^*} < H_{K=k^*}$ and $H_{K=k^*} \approx H_{K=k^*+j}$ for $j > 0$. 

For datasets that exhibit natural groupings, it is sometimes reasonable to suppose that the datapoints within these groups follow some prescribed distribution (like Gaussians, as assumed above).  It is harder, however, to anticipate how these group centroids might be distributed, and so a scaling analysis of BSS based on underlying data distributions, as performed for WSS, is not possible in general.   Generically, though, we know that Euclidean squared distances scale linearly with dimension, and so we can write ${\rm BSS}_{K} = \alpha + d\beta$, for $\alpha,\beta$ determined by the actual distribution of cluster and data centroids.  From above, we know that ${\rm WSS}_{K=k^*} \propto d$ and so we expect that ${\rm H}_{K=k^*} \rightarrow {\rm const.}$ as $d \rightarrow \infty$.  In practice, the Hartigan index becomes constant quickly, when $d \sim \mathcal{O}(10)$.  Because both ${\rm BSS}_{K=k^*-1}$ and ${\rm WSS}_{K=k^*-1}$ also scale linearly with $d$, the sensitivity of the Hartigan index, $s_{\rm H} = \lvert {\rm H}_{K=k^*-1} - {\rm H}_{K=k^*}\rvert \rightarrow {\rm const.}$ as the dimensionality increases\footnote{ As we will see in the next section, by transforming the Hartigan index (and other knee- and elbow-type indices) into optimization problems, for some the sensitivity improves with dimension. We will discuss why in the next section.}

In this section, we have analytically examined how the sensitivities of two relative validity indices scale with increasing dimension for the case of well-separated Gaussian clusters.  This analysis was simply to motivate, via example, our notion of sensitivity and to contrast it with the scaling behavior of indices evaluated at $K=k^*$.  We also observed that the sensitivity scales according to which base quantities are included in the index ({\it e.g.} within-cluster vs between-cluster) and how they are functionally combined.   There are many dozens of other relative validity criteria in the literature of wildly different origin, but all are ultimately defined in terms of quantities relevant on one of three scales: within-cluster (W), between-cluster (B), and full dataset (D).  For example, Ball-Hall is based on within-cluster distances (W-type), and Hartigan is defined in terms of both within- and between-cluster quantities (WB-type).  An example of an index that incorporates a full dataset measure is the {\it R-squared} index,
\begin{equation}
{\rm RS} = 1 - \frac{\sum_k^K\sum_{x_i \in C_k} \lvert \lvert x_i - \overline{x}_k\rvert \rvert^2}{\sum_{x_i \in \mathcal{D}} \lvert \lvert x_i - \overline{x}\rvert \rvert ^2} = 1 - \frac{{\rm WSS}}{{\rm TSS}}
\end{equation}
where the denominator is the sum of squared distances between each data point and the centroid of the full dataset.  The RS measure is also a function of within-cluster distances and so is of type-WD.  

In the next section, we examine the scaling behavior of sensitivities of 24 relative validity criteria under a few different data schemes, including well-separated univariate Gaussians as well more noisy distributions.  In addition to studying how the various criteria scale according to data scheme, we are also interested in learning whether the type of index ({\it e.g.} W vs WD) influences scaling behavior in any discernible way.

\begin{figure*}[ht]
\centering
\includegraphics[width=6in]{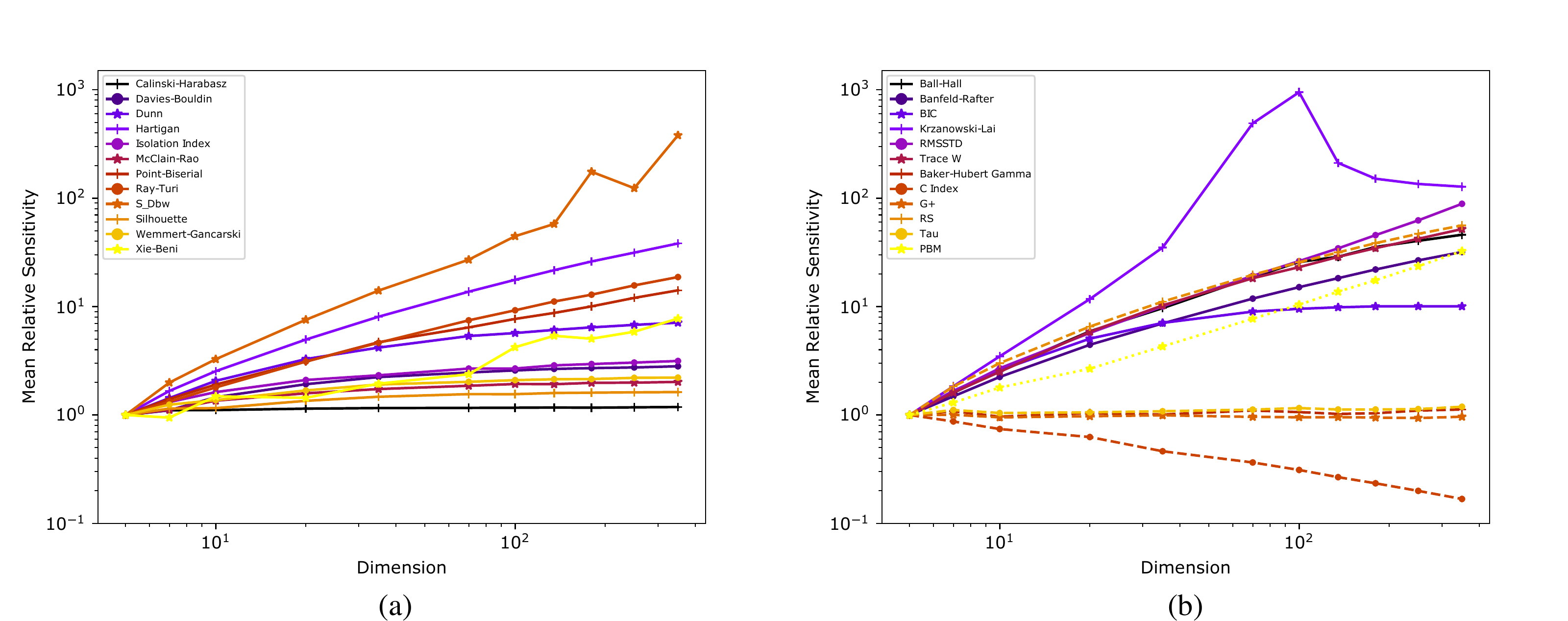}
\caption{\footnotesize{Mean relative sensitivity of (a) WB-type indices, and (b) W-, WD-, and WBD-type indices. Sensitivities less than one indicate a loss of sensitivity relative to that achieved at $d=5$.}}
\label{s_uni}
\end{figure*}
\section{Experiments}
The sensitivities of 24 relative validity criteria from the recent literature \cite{Liu2010,Vendramin,Tomasev,Desgraupes} (listed in Table \ref{T} and defined in the Appendix) are evaluated in this section for five different data schemes: well-separated, univariate Gaussian clusters, well-separated multivariate Gaussian clusters, well-separated univariate Gaussian clusters immersed in uniform noise, and well-separated univariate Gaussian clusters with some percentage of features (10\% or 50\%) uniformly distributed (to simulate un-informative, or irrelevant, features). These latter two data distributions are equivalent to Gaussian data with highly-colored noise.  We vary the dimension of the space over the range $d \in [5,7,10,20,35,70,100,135,180,250,350]$, avoiding dimensions fewer than 5 because we find that uniformly distributed clusters tend to overlap\footnote{It is this overlapping of clusters in lower dimensions that we believe is responsible for the apparent improvement at higher-dimensions reported in \cite{} for certain indices like the silhouette score}.  We present here results for $k^*=5$, though like \cite{Tomasev} we observe that sensitivity as a function of dimension is largely insensitive to the value of $k^*$

Each index is tested against 100 realizations of each data scheme, and the sensitivities at each dimension are averaged over the 100 realizations.  We report this average sensitivity relative to that at $d=5$ (the lowest dimension that we test), $\overline{s}/\overline{s}_{d=5}$.   For each dimension and data realization, $k$-means clustering was used to partition the data across a small interval of $k$-values centered on $k^*$.  We note that $k$-means correctly partitioned when $K=k^*$ for all data realizations in all dimensions tested, consistent with the dimensionality study conducted in \cite{Sieranoja}.
\begin{table}[]
\begin{tabular}{llll}
Index & Optimum  & Type  & Reference  \\
\hline \\
Baker-Hubert $\Gamma$ & Max  & W & \cite{Baker}  \\
Ball-Hall & Elbow  & W & \cite{Ball}   \\
Banfield-Raftery &Elbow  &W  &  \cite{Banfield}  \\
BIC &Elbow  &W  & \cite{Pelleg}   \\
C Index &Min  &WD  & \cite{Hubert}   \\
Cali\'{n}ski-Harabasz&Max  &WB  & \cite{Calinski}   \\
Davies-Bouldin&Min  &WB  &  \cite{Davies}  \\
Dunn&Max  &WB  & \cite{Dunn}   \\
G+&Min  &WD  &  \cite{Rohlf}  \\
Isolation Index&Knee  &WB  &\cite{Frederix}    \\
Krzanowski-Lai&Max  &W  &  \cite{Krzanowski}  \\
Hartigan&Knee  &WB  &  \cite{Hartigan}  \\
McClain-Rao&Elbow  &WB  &  \cite{McClain}  \\
PBM&Max  &WBD  &  \cite{Pakhira}  \\
Point-Biserial&Max  &WB  &  \cite{Milligan}  \\
RMSSTD&Elbow  &W  & \cite{Halkidi}   \\
RS&Knee  &WD  & \cite{Halkidi}   \\
Ray-Turi&Elbow  &WB  & \cite{Ray}   \\
{\it S\_Dbw}&Elbow &WB  & \cite{Halkidi2001}   \\
Silhouette&Max &WB  & \cite{Rousseeuw}   \\
$\tau$ &Max &WD  & \cite{Milligan}   \\
Trace W&Elbow &W  &  \cite{Edwards}  \\
Wemmert-Gan\c{c}arski&Max &WB  &  \cite{Desgraupes}  \\
Xie-Beni&Elbow &WB  & \cite{Xie}   
\end{tabular}
\caption{\footnotesize{Relative validity criteria analyzed in this study.}}
\label{T}
\end{table}

All knee- and elbow-type indices are converted into optimization problems by applying the transformation \cite{Vendramin},
\begin{equation}
m'(K) = \left\lvert \frac{m(K-1) - m(K)}{m(K) - m(K+1)}\right\rvert.
\end{equation}
We mentioned in the last section that the sensitivity of the knee-type Hartigan index improves with dimension after applying the above difference transformation.   This is because the points on the ``thigh'' (with $K \geq k*$) flatten-out in higher dimensions and so the difference $m(K) - m(K+1)$ becomes small, driving up the value of $m'(K)$.
\subsection{Univariate Gaussian clusters}
We first consider data with $k^*$ univariate Gaussian clusters.  The centroid of each cluster is selected uniformly from the range $[-10,10]$, and the variance of each cluster is fixed at unity.  The number of points in each cluster are selected from a normal distribution with $\mu = 200$ and $\sigma = 10$.

The sensitivities of the indices are plotted as a function of dimension in Figure \ref{s_uni}.  The 12 WB-type indices are plotted separately (a), and the 6 W-type, 5 WD-type (dashed), and 1 WBD-type (dotted) indices are plotted together in (b).  
\begin{figure*}[ht]
\centering
\includegraphics[width=6in]{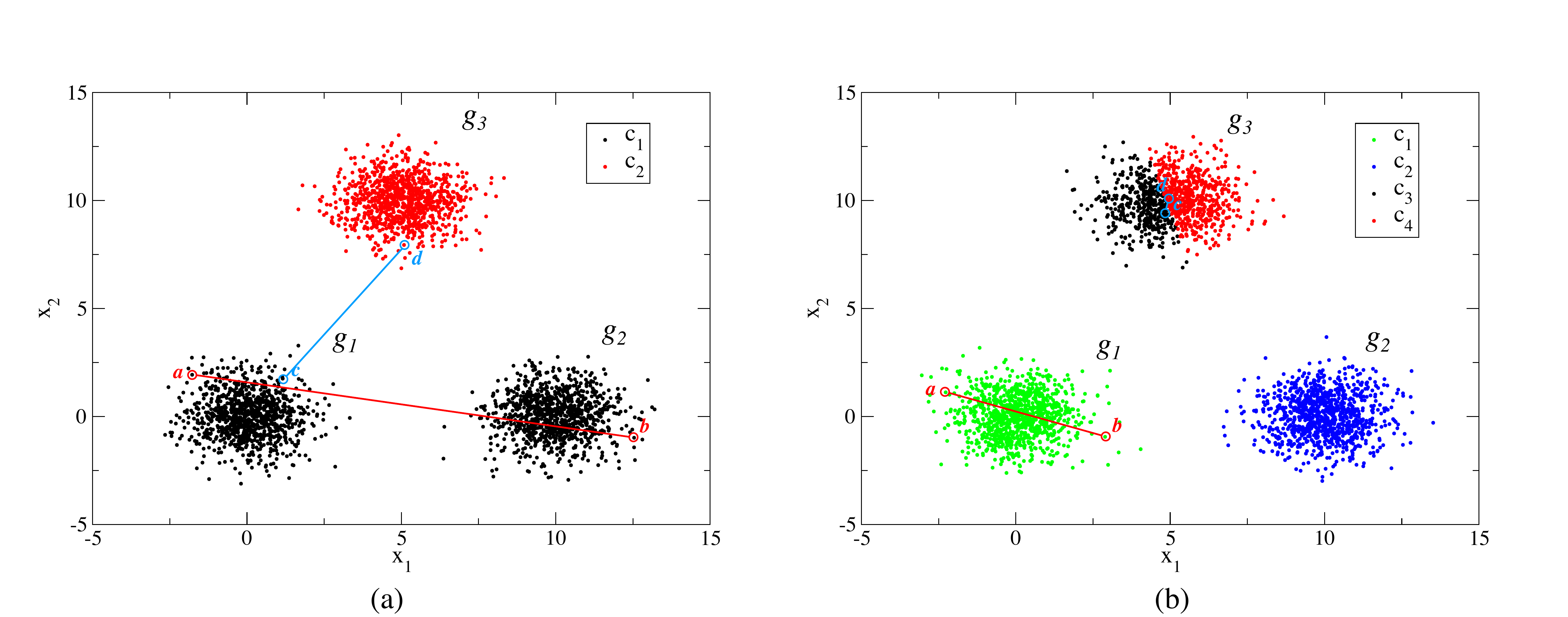}
\caption{\footnotesize{(a) Case $K = k^*-1$: When a single cluster spans two natural groupings, two points within this cluster $(a,b)$ can have a greater separation than two points $(c,d)$ in different clusters. (b) Case $K = k^*+1$: When a single natural grouping is split by two clusters, two points within this grouping but in different clusters $(c,d)$ can be closer together than two points $(a,b)$ within the same cluster spanning a single natural grouping.}}
\label{sminus}
\end{figure*}
All the WB-type indices have improved sensitivity as dimensionality increases for well-separated, Gaussian clusters. Of those indices in Figure \ref{s_uni} (b), only the C index noticeably degrades as dimensionality increases.  All of the other WD-type indices (except for RS) have approximately stable sensitivities, that is, they have $\overline{s}/\overline{s}_{d=5} \approx 1$.  Stable indices are {\it scale free}, with the desirable property that we know precisely how they will behave in higher dimensions.  They are of considerable interest so warrant further investigation.  This group of WD-type indices: Baker-Hubert $\Gamma$, G+, and $\tau$, are functions of the quantities $s^+$ and/or $s^-$, where $s^+$ is the number of {\it concordant}, and $s^-$ the number of {\it discordant} pairs of points.  Two data points $a$ and $b$ within the same cluster form a concordant pair if they are closer together than any two points $c$ and $d$ not within the same cluster; if $a$ and $b$ do not form a concordant pair, then there exist points $c$ and $d$ that form a discordant pair.  

To understand the stability of these quantities' sensitivities, consider, for example, the quantity $s^-$: when fewer than $k^*$ clusters are used to partition the data, at least one cluster will contain at least two natural groupings so that $a$ and $b$ might be quite far apart.  Referring to Figure \ref{sminus} (a), a particular pair of points $a$ and $b$ within the same cluster spanning the two groups $g_1$ and $g_2$ will be farther apart than a great number of pairs $c$ and $d$ in different clusters, so $s^-$ is large when $K < k^*$.  Likewise, referring to Figure \ref{sminus} (b), when $K > k^*$, $s^-$ is large because in this case at least one natural grouping is split among at least two clusters, so that points $c$ and $d$ within such a split grouping (but within different clusters) will be relatively close together in comparison to points $a$ and $b$ on opposite sides of some other cluster.  The minimum of $s^-$ across a range of $K$ therefore picks out $k^*$, and we argue now that the sensitivity of $s^-$ is largely independent of dimension. 

First, we argue that $s^{-}_{K=k^*-1} > s^{-}_{K=k^*+1}$.  When $K = k^*-1$, of order ${n_c \choose 2}$ pairs of points $(a,b)$ (from a single cluster spanning two natural groupings) will have a separation greater than the closest pair of points $c$ and $d$ in different clusters.  When $K = k^*+1$, fewer than $\frac{1}{2}{n_c \choose 2}$ pairs of points $(c,d)$ in the different clusters spanning a single natural grouping will be closer together than pairs of points $a$ and $b$ in some other cluster, as in Figure \ref{sminus} (b).  There is then an upper bound, $s^{-}_{K = k^*-1} > 2s^{-}_{K = k^*+1}$, which is actually very liberally satisfied in practice.  All of this is to conclude that $s = \lvert s^{-}_{K=k^*} - s^{-}_{K=k^*+1}\rvert$.  If we suppose that $s^{-}_{K = k^*} = 0$ (this is the case seen empirically for well-separated clusters), it remains to then show that $s^{-}_{K=k^*+1}$ is independent of dimension.

The focus is on how the distances between pairs of points $(c,d)$ in different clusters of a split grouping compare with the distances between points $(a,b)$ within the same grouping (as in Figure \ref{sminus} (b)).  Assuming for simplicity that all clusters are of equal size, we can focus on a single Gaussian cluster and study the following problem: for each pair of points $c$ and $d$ existing in the separate regions of a bisected version of this cluster, how many pairs of points $a$ and $b$ existing anywhere within the complete cluster have a greater separation, $D_{ab} > D_{cd}$?  The distribution of Euclidean distances between Gaussian-random points has been studied in \cite{Thirey}, where it is found that as $d \rightarrow \infty$, this distribution resembles a shifted Gaussian with mean, $\mu$, growing with $d$ (in fact, this distribution becomes very nearly normal for relatively low dimensions, $d \gtrsim 10$).  This shift away from $\mu = 0$ is due to the fact that distances tend to grow as the dimensionality of the space increases: quite simply, there is more space to move around in.  Meanwhile, the variance of the shifted Gaussians is approximately independent of $d$, with the result that the number of points with a separation greater then a fixed proportion, $p$,  of the mean separation, $D_{ab} > p\mu$, is independent of dimension.  This is relevant to our question: the number of pairs $(c,d)$ with separation $p\mu$ is constant as dimensionality increases, as is the number of pairs $(a,b)$ with $D_{ab} > D_{cd}$.  The number of all such pairs is $s^-_{K=k^*+1}$.  The quantity $s^-$ is therefore independent of dimension.   
\begin{figure*}[ht]
\centering
\includegraphics[width=6in]{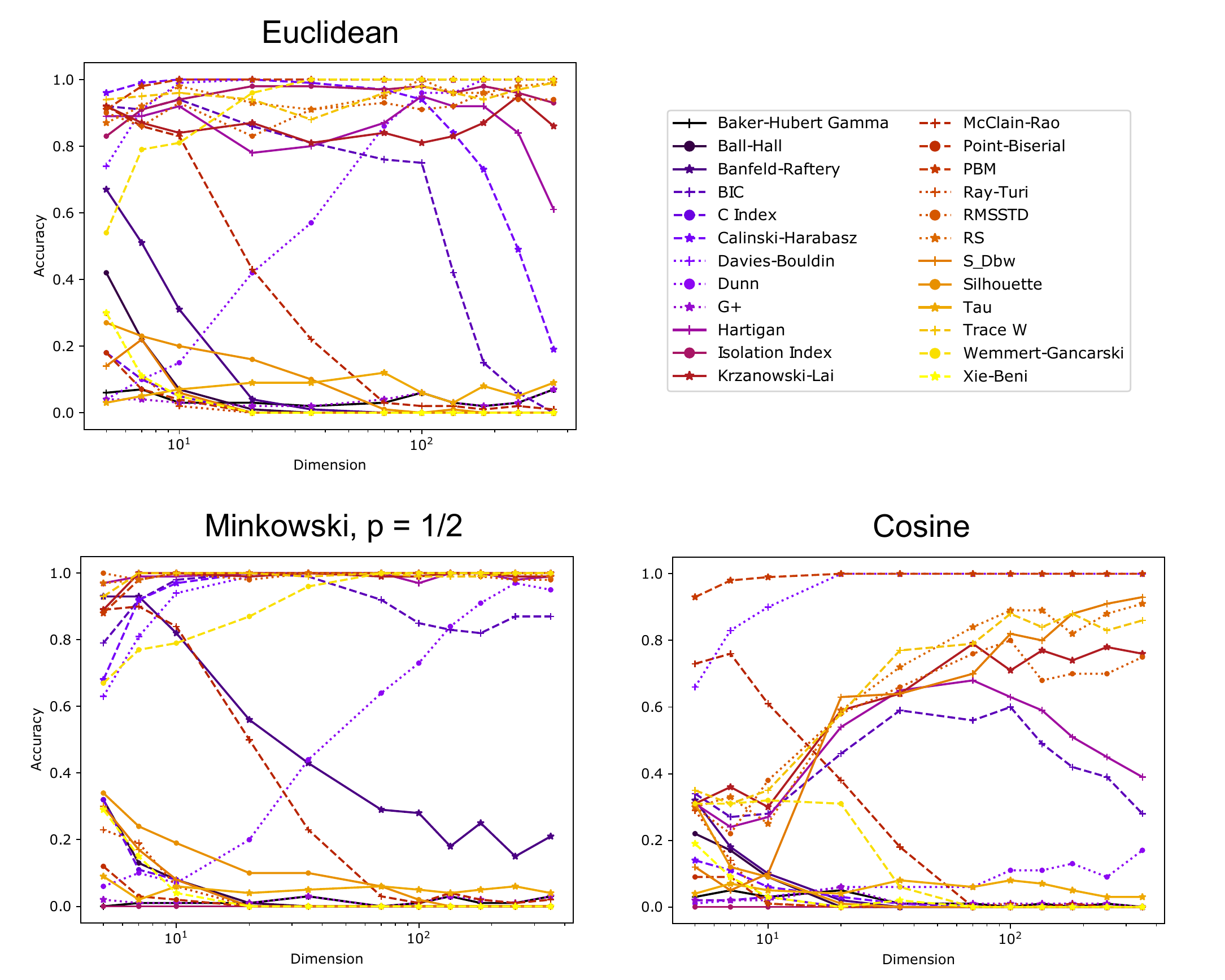}
\caption{\footnotesize{Accuracy of indices as a function of dimension for three different distance metrics: Euclidean, Minkowski with $p=1/2$, and cosine similarity, when tested against Gaussian clusters immersed in uniform noise.}}
\label{acc}
\end{figure*}

As with the Ball-Hall and Hartigan indices analyzed in the last section, we see that statistical arguments can be used to analyze the sensitivity of indices based on the number of concordant and/or discordant pairs, like the Baker-Hubert $\Gamma$, G+, and $\tau$.  This kind of analysis is useful also for designing validity indices with predictably good high-dimensional behavior.   

We also performed the above testing for well-separated, multivariate Gaussian clusters, with data prepared as above but with the standard deviation selected randomly from the uniform interval $[0.5,1.5]$ for each dimension.  The results are qualitatively very similar to the above, again with all indices except the C index either improving or stable with increasing dimension.

In summary, we can conclude from these results that the great majority of relative validity criteria see {\it improved} sensitivity in high-dimensional data spaces with well-separated natural groupings.  
\subsection{Univariate Gaussian clusters with noise}
The above case is an idealized scenario, as few real-world datasets are comprised of clean, well-separated groupings.  To create noisy data, we generate $k^*$ univariate Gaussian clusters as above and then add uniform noise over the interval $[-10,10]$ in each dimension.  This is a kind of white noise, present with equal probability in each dimension.  Since $k$-means will assign each noise point to a cluster, when $k=k^*$ the clusters will be strongly Gaussian with noisy points occurring as outliers in at least one dimension. The noise level is quantified as a percentage of the number of points in the dataset: since our clusters have on average 200 points, a 1\% noise level works out to about 2 noisy points per cluster.  Though a seemingly tiny amount of noise, we find that some indices are badly affected at the 1\% level.  In Figure \ref{acc} (top left) the {\it accuracy} of each index---the proportion of data instantiations in which the index correctly picked out $k^*$---is plotted as a function of dimension.   Several indices perform poorly, with low accuracy already at $d=5$ that degrades to near 0\% in higher dimensions.  
Meanwhile, a small collection of indices perform well for all $d$: Davies-Bouldin, Krzanowski-Lai, PBM, RMSSTD, RS, and Trace $W$, each maintaining an accuracy above 80\% across the range of tested dimensions, showing that noisy data does not compound the curse of dimensionality for these indices. 

More interesting are the indices that perform relatively well in low dimensions but degrade as $d$ increases: Banfield-Raftery and McClain-Rao are two such indices that degrade rather quickly, whereas BIC, Cali\'{n}ski-Harabasz, and the Hartigan indices degrade more abruptly only when $d$ surpasses 100 or so.  From Figure \ref{dist}, we see that uniform data distributions more strongly exhibit the issues of distance concentration and hub formation, suggesting that the uniform noise introduced in this data scheme might be contributing to the reduced accuracy of these indices. To test this, one might replace the Euclidean distance used in these indices with a distance metric that has better high dimensional performance.  Several authors have studied and argued for the use of Minkowski metrics with fractional powers in high-dimensional spaces,
\begin{equation}
L_p(x,y) = \left(\sum_{i=1}^d \lvert \lvert x_i - y_i \rvert \rvert ^p\right)^{1/p}.
\end{equation}
This metric generalizes the familiar Euclidean-squared, or $L_2$ norm, to arbitrary powers.  There is some evidence \cite{Aggarwal2001,Aggarwal22001,Doherty2004,Verleysen05,Hsu,Jayaram} that norms with $0 < p < 1$ are more resistant to distance concentration in high dimensions than the $L_2$ norm.  

There is also the {\it cosine similarity},
\begin{equation}
d_C(x,y) = \cos(\theta) = \frac{{\mathbf x}\cdot {\mathbf y}}{\lvert \lvert {\mathbf x} \rvert \rvert\, \lvert \lvert {\mathbf y} \rvert \rvert},
\end{equation}
which associates the closeness of two points with the degree of co-directionality of their vectors.  It has found use in high dimensional generalizations of $k$-means (via the spherical $k$-means algorithm \cite{Hornik}), in which points are projected onto the surface of a hypersphere via the cosine similarity, and then partitioned on this surface using conventional $k$-means.  

To determine whether these metrics might improve cluster validation in high dimensional spaces in the presence of uniform noise, each index in Table \ref{T} was redefined in terms of the Minkowski metric with $p=1/2$ and the cosine similarity, and the above experiment repeated.  Results are shown in Figure \ref{acc} (bottom).  The Minkowski metric offers some improvement, particularly by recovering the performance of the BIC, Cali\'{n}ski-Harabasz, and Hartigan indices that are flagging in high dimensions with Euclidean distance.  Meanwhile, the cosine similarity generally under-performs Euclidean distance at low dimension, but supports a fairly general improvement with $d$ for a number of indices.      

\begin{figure*}[ht]
\centering
\includegraphics[width=6in]{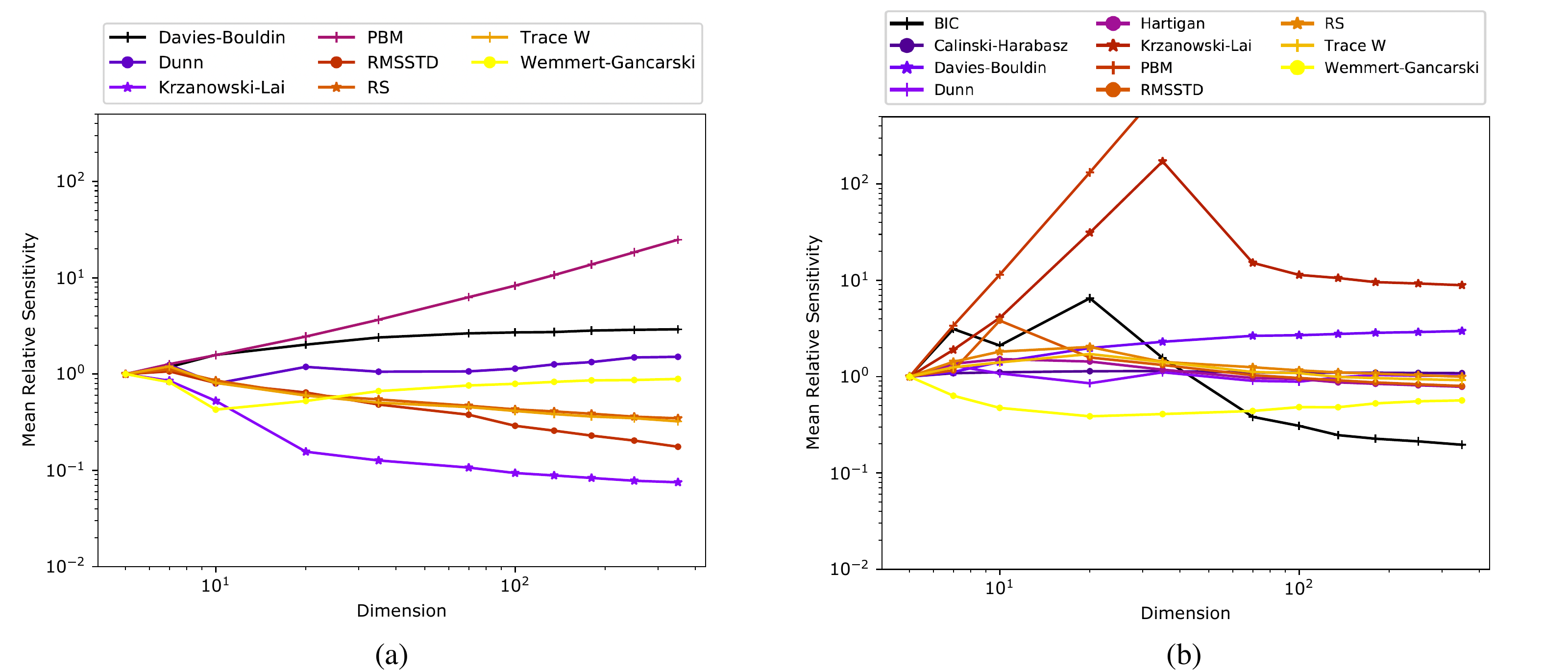}
\caption{\footnotesize{Mean relative sensitivity of indices from Figure \ref{s_noise} with greater than 80\% accuracy with (a) Euclidean distance and (b) Minkowski.}}
\label{s_noise}
\end{figure*}
Focusing only on those indices with accuracies surpassing 80\% across the full range of tested dimensions, we next examine their sensitivities in Figure \ref{s_noise}.  While noise does not affect the accuracy of these indices, it does attenuate their sensitivities, causing some based on Euclidean distance to degrade with increasing dimension.  Use of Minkowski distance with $p=1/2$ tends to ease this difficulty, resulting in most indices performing stably across dimensions. 

\begin{figure*}[ht]
\centering
\includegraphics[width=6in]{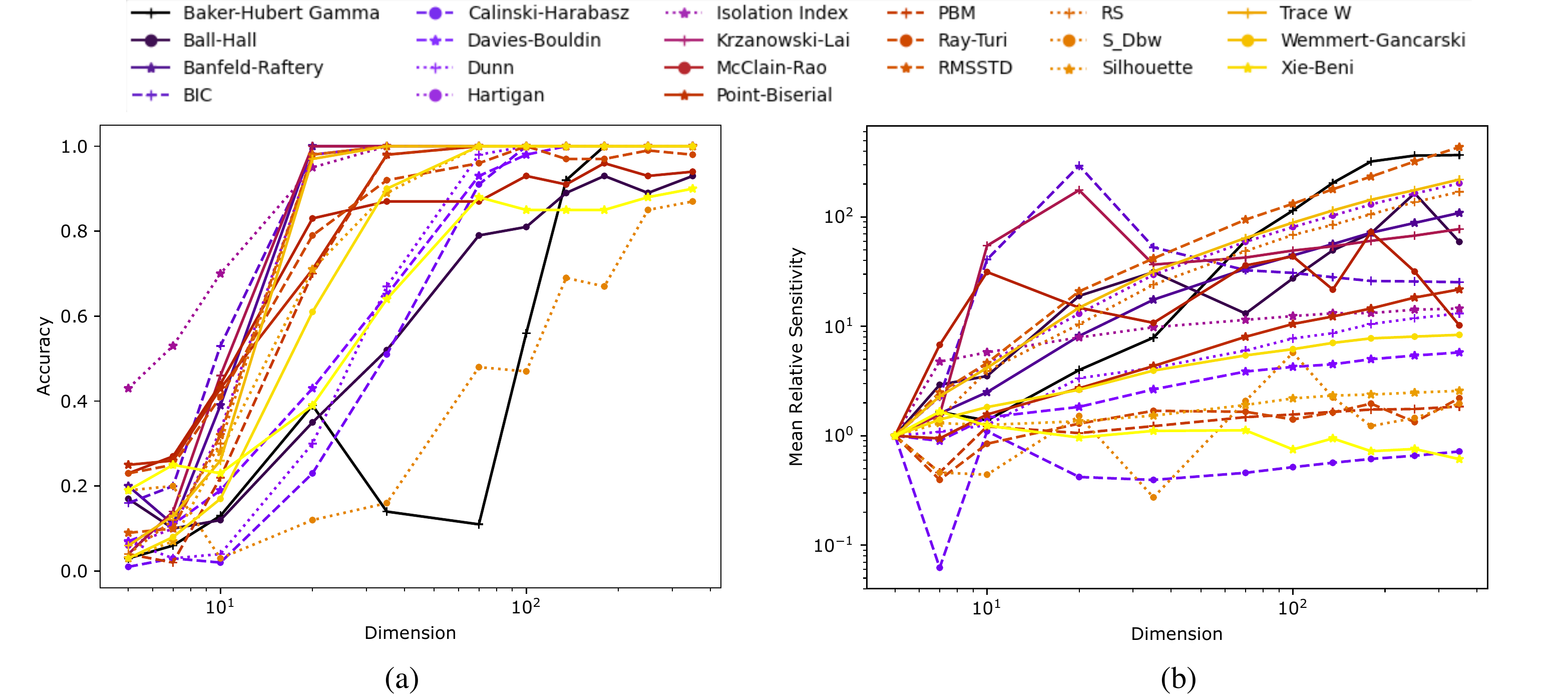}
\caption{\footnotesize{(a) Accuracy and (b) mean relative sensitivity of Euclidean indices when tested against Gaussian clusters with 50\% irrelevant features.}}
\label{s_acc_5}
\end{figure*}
As the noise level is increased, performance degrades; for example, at a 10\% noise level, all indices drop to below around 20\% accuracy across all dimensions.  The presence of noise is thus a direct challenge to validity indices even in low dimensions, and is not compounded by increasing dimensionality.  In summary, we find that for low levels of noise, certain Euclidean-based indices perform accurately in all tested dimensions, and the fractional Minkowski metric offers both improved accuracy and sensitivity for some of these indices.
\subsection{Univariate Gaussian clusters with irrelevant features}
It sometimes occurs that not all features follow patterns that are useful for modeling.  In the context of clustering problems, these features tend to follow uniform random or other non-central distributions, in contrast to relevant features that tend to be more centrally distributed.  The result is that the data do not form well-defined clusters in the irrelevant dimensions.  In this section, we study the effect of irrelevant features on cluster validation in high-dimensional spaces by considering two data schemes, in which 10\% and 50\% of features are irrelevant.

To create these datasets, for each cluster we randomly select a percentage of features (10\% or 50\%) and draw the values of these features for each data point from the uniform range $[-10,10]$.  All other features are deemed relevant and drawn from a univariate Gaussian distribution as above.  The results for validation indices with Euclidean distance are provided in Figure \ref{s_acc_5} for 50\% irrelevant features.  Remarkably, accuracies are generally low in low-dimensional feature spaces (a) and, without exception, improve with increasing dimension.  Relative sensitivities are also generally increasing functions of $d$, except for the Cali\'{n}ski-Harabasz and Xie-Beni indices (b).  Results for 10\% irrelevant features are qualitatively similar, with slightly better performance and sensitivity. A small set of indices: C index, G+, and $\tau$, could not be analyzed because accuracy was zero at $d=5$, which prevented an evaluation of $\overline{s}_{d=5}$ and hence relative sensitivity, are not defined.  

\begin{figure}[ht]
\centering
\includegraphics[width=3in]{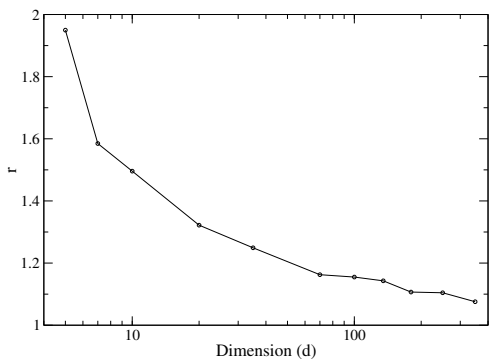}
\caption{\footnotesize{Ratio of Eq. \ref{ratio} as a function of dimension.  The decrease reveals that distortion of clusters due to irrelevant features is mitigated by distance concentration in high dimensions.}}
\label{r}
\end{figure}

This improvement of accuracy with dimension can be understood as follows.  First, notice that irrelevant features tend to deform clusters uniformly along the corresponding dimension.  For example, consider a cluster in 3 dimensions that is Gaussian-distributed in two dimensions and uniformly distributed along one (and so has one irrelevant dimension).  The resulting distribution is a uniform cylinder with a Gaussian cross-section.  Importantly, the effect of irrelevant features is not to create random noise throughout the dataset, but to uniformly stretch and deform clusters. In higher dimensions, the effect of distance concentration comes into play: it acts to {\it de-emphasize} the stretched dimensions relative to the relevant features resulting in a more localized cluster.  To see this quantitatively, we define the ratio 
\begin{equation}
r = \frac{\max_i \lvert \lvert x_i - \overline{x} \rvert \rvert}{\frac{1}{N}\sum_i \lvert \lvert x_i - \overline{x} \rvert \rvert},
\label{ratio}
\end{equation}
which is the maximum distance from the cluster centroid relative to the average distance.  As the dimension of the space grows, this ratio tends to unity, as shown in Figure \ref{r} for a single cluster with half of its features irrelevant. 
As clusters become more localized in higher dimensions, validity measures are able to more accurately resolve them and accuracy improves.  Use of the fractional Minkowski metric, shown to be particularly useful for defining nearest neighbors in data with colored noise \cite{Francois,Francois2007}, offers only marginal improvements to accuracy and sensitivity. 

\section{Conclusions}
In this work, the accuracies and sensitivities of a set of 24 popular cluster validation measures were studied as a function of feature space dimensionality for a variety of data models.  For well-separated, multivariate Gaussian clusters, we find that all but one index have either improving or stable sensitivity as the dimensionality of the space is increased.  When uniform noise is introduced, performance degrades for some indices in higher dimensions but this can be corrected by replacing Euclidean distance with fractional Minkowski metric with $p=1/2$.  Finally, indices also perform well for data models with up to 50\% irrelevant features, with sensitivities and accuracies that improve with increasing dimension.  These results paint an optimistic picture: {\it cluster validation using relative indices is not challenged by distance concentration in high dimensional spaces} for Gaussian clusters with white or colored noise.  For well-separated Gaussian clusters with or without colored noise (irrelevant features) index sensitivities actually improve with increasing dimension.

While simulated data is no substitute for real-world datasets, it is the only way to systematically isolate and explore the effect of dimensionality on cluster validation.  The data schemes considered in this analysis are meant to address both clean and noisy clustering problems; however, the success of relative criteria ultimately depends on the nature and qualities of the dataset.  Due to the fact that virtually all tested indices show success in high dimensions, there is considerable room to select indices appropriate to a given problem.  

This analysis has considered relative criteria, as opposed to validation methods based on absolute criteria, like g-means or dip-means, which select partitions that are Gaussian and unimodal, respectively.  These two methods are known to degrade considerably in higher dimensions \cite{Adolfsson}, making the success of relative criteria all the more critical.  A systematic analysis like that done here for relative criteria might also be conducted for different absolute criteria to further our understanding of their performance. 

In conclusion, it is hoped that this study strengthens our confidence in the face of the curse of dimensionality, revealing that quantities based on the Euclidean norm can still be useful in high dimensional spaces if they are defined to scale properly.  Happily, many of the well-known criteria are so defined, and therefore find continued applicability even, and particularly for, high-dimensional problems.


\section{Appendix}
In this section we provide definitions and descriptions of the 24 internal validity indices tested in this work.  

\subsection{Baker-Hubert $\Gamma$}
The $\Gamma$ index measures the correlation between the pairwise distances between points, and whether these pairs belong to the same cluster or not.  Defining $s^+$ as the number of times the distance between two points belonging to the same cluster is strictly smaller than the distance between two points not belonging to the same cluster (the number of {\it concordant} points) and $s^-$ as the number of times the distance between two points not belonging to the same cluster is strictly smaller than the distance between two points belonging to the same cluster (the number of {\it discordant} points), the $\Gamma$ index is defined,
\begin{equation}
\Gamma = \frac{s^+ - s^-}{s^+ + s^-}.
\end{equation}
\subsection{Ball-Hall}
The Ball-Hall index is the mean dispersion averaged over $K$ clusters, 
\begin{equation}
{\rm BH} = \frac{1}{K} \sum^{K}_{k=1} \overline{\rm WSS}_k.
\end{equation}
\subsection{Banfield-Raftery}
Similar to the Ball-Hall index, the Banfield-Raftery index is the weighted sum,
\begin{equation}
{\rm BR} = \sum^{K}_{k=1} n_k \log (\overline{\rm WSS}_k).
\end{equation}
\subsection{BIC}
The Bayesian information criterion (BIC) measures the optimal number of parameters needed by a model to describe a dataset. It was adapted to the clustering problem for the $X$-means algorithm \cite{Pelleg} and is written,
\begin{eqnarray}
{\rm BIC} &=& c + \sum_{k=1}^K \left[n_k \log \frac{n_k}{N} - \frac{(n_k -1)d}{2} \right.  \nonumber \\
&&\left. - \frac{n_k d}{2}\log \left(\frac{2\pi  n_k d}{N-K}\sum_{k=1}^K \overline{\rm WSS}_k\right)\right],
\end{eqnarray}
where $c = K(d+1)\log(N)/2$ is a constant, $N$ is the number data points, and $d$ is the number of dimensions.
\subsection{C index}
In each cluster $C_k$, there are $n_k(n_k -1)/2$ pairs of points and there are thus $N_W = \sum_{k=1}^K n_k(n_k -1)/2$ pairs of points in the same cluster across the full dataset.   Defining $S_W$ as the sum of the distances between all pairs of point belonging to the same cluster, $S_{\rm min}$ as the sum of the $N_W$ closest pairs of points in the full dataset, and $S_{\rm max}$ as the sum of the $N_W$ most-distant pairs of points in the full dataset, the C index is 
\begin{equation}
C = \frac{S_W - S_{\rm min}}{S_{\rm max} - S_{\rm min}}.
\end{equation}
\subsection{Cali\'{n}ski-Harabasz}
With the between-cluster dispersion defined in Eq. \ref{bss}, the Cali\'{n}ski-Harabasz index is defined
\begin{equation}
{\rm CH} = \frac{N-K}{K-1}\frac{{\rm BSS}}{{\rm WSS}},
\end{equation}
where WSS is the pooled within-cluster sum of squares, ${\rm WSS} = \sum_k {\rm WSS}_k$.
\subsection{Davies-Bouldin}
For each cluster $C_k$, define the mean distance of the cluster points to the cluster barycenter,
\begin{equation}
\delta_k = \frac{1}{n_k} \sum_{x_i \in C_k} \lvert \lvert x_i - \overline{x}_k\rvert \rvert.
\end{equation}
The distance between barycenters of clusters $C_k$ and $C_k'$ is $\Delta_{k k'} = \lvert \lvert \overline{x}_k - \overline{x}_{k'} \rvert \rvert$.  The Davies-Bouldin index concerns the maximum of the ratio $\frac{\delta_k + \delta_{k'}}{\Delta_{k k'}}$ for each $k$ (with $k' \neq k$) averaged over all clusters,
\begin{equation}
{\rm DB} = \frac{1}{K} \sum_{k=1}^K \max_{k' \neq k} \left(\frac{\delta_k + \delta_{k'}}{\Delta_{k k'}}\right).
\end{equation}
\subsection{Dunn}
These are really a family of measures known as the {\it generalized Dunn indices} based on the ratio of some measure of between-cluster ($\delta$)  and within-cluster ($\Delta$) distances,
\begin{equation}
{\rm GDI} = \frac{\min_{k\neq k'} \delta (C_k,C_{k'})}{\max_k \Delta (C_k)}.
\end{equation}
In this study we define $\Delta(C_k) = \max_{x_i,x_j \in C_k} \lvert \lvert x_i - x_j \rvert \rvert$ to be the maximum distance between two pairs of points in cluster $C_k$ and 
\begin{equation}
\delta(C_k,C_{k'}) = \min_{x_i \in C_k, x_j \in C_{k'}} \lvert \lvert x_i - x_j \rvert \rvert,
\label{gdi}
\end{equation}
the distance between clusters $C_k$ and $C_{k'}$ defined to be the minimum distance between a point $x_i$ in $C_k$ and a point $x_j$ in $C_{k'}$.  This is known as {\it single linkage}. 
\subsection{G+}
Using the definition of  $s^-$ defined above for the Baker-Hubert $\Gamma$ index,
\begin{equation}
{\rm G}+ = \frac{2s^-}{N_T(N_T-1)},
\end{equation}
where $N_T$ is the total number of distinct pairs of points in the full dataset, $N_T = N(N-1)/2$.
\subsection{Isolation index}
Denoting the set of the $p$ nearest neighbors of a point $x_i$ in cluster $C_k$ by $N_{i,p}$, the quantity $\delta_{i,p}$ is the number of points $x_j \in N_{i,p}$ that are {\it not} in $C_k$.  The isolation index is then defined,
\begin{equation}
{\it IS} = \frac{1}{N}\sum_{x_i \in \mathcal{D}} \left(1 - \frac{\delta_{i,p}}{p}\right),
\end{equation}
where $\mathcal{D}$ denotes the set of all data points.
\subsection{Krzanowski-Lai}
An evaluation of the Krzanowski-Lai index for $K$ clusters requires three partitionings, at $K-1$, $K$, and $K+1$.  Defining ${\rm diff}_K = (K-1)^{2/d}{\rm WSS}_{K-1} - K^{2/d}{\rm WSS}_{K}$, we have
\begin{equation}
{\rm KL} = \frac{\lvert{\rm diff}_K\rvert}{\lvert{\rm diff}_{K+1}\rvert}.
\end{equation}
In principal the value of this index depends on the clustering algorithm selected to perform the partitioning;  in this work we chose $k$-means.
\subsection{Hartigan}
This index is the logarithm of ratio of the between-cluster dispersion and the pooled within-cluster dispersion, both defined earlier,
\begin{equation}
{\rm H} = \log \frac{{\rm BSS}}{{\rm WSS}}.
\end{equation}
\subsection{McClain-Rao}
As with $S_W$, the sum of the distances between all pairs of points belonging to the same cluster (defined for the C index above), we can denote the sum of the distances between all pairs of points not belonging to the same cluster as $S_B$.  There are $N_B = N(N-1)/2 - N_W$ such pairs.  The McClain-Rao index is the ratio,
\begin{equation}
{\rm MR} = \frac{N_B}{N_W}\frac{S_W}{S_B}.
\end{equation}
\subsection{PBM}
Let $D_B = \max_{k < k'} \lvert \lvert \overline{x}_k - \overline{x}_{k'} \rvert \rvert$ be the maximum distance between any two cluster barycenters, $E_W = \sum_{k=1}^K \sum_{x_i \in C_k} \lvert \lvert x_i - \overline{x}_k\rvert \vert$ be the sum of the distances of the points of each cluster to their barycenter, and $E_T = \sum_{x_i \in \mathcal{D}} \lvert \lvert x_i - \overline{x}\rvert \vert$ be the sum of the distances of all points to the data barycenter.  Then,
\begin{equation}
{\rm PBM} = \left(\frac{E_T D_B}{K E_W}\right)^2.
\end{equation}
\subsection{Point-biserial}
This index measures the correlation between the distances between pairs of points and whether the points belong to the same cluster.  In terms of quantities defined above,
\begin{equation}
{\rm PBI} = \left(\frac{S_W}{N_W} - \frac{S_B}{N_B}\right)\frac{\sqrt{N_W N_B}}{N_T}.
\end{equation}
\subsection{RMSSTD}
The root mean square standard deviation (RMSSTD) is defined
\begin{equation}
{\rm RMSSTD} = \sqrt{\frac{{\rm WSS}}{d\sum_{k=1}^K (n_k - 1)}}.
\end{equation}
\subsection{RS}
The R-squared (RS) index is defined
\begin{equation}
{\rm RS} = 1 - \frac{{\rm WSS}}{{\rm TSS}},
\end{equation}
where ${\rm TSS} = \sum_{x_i \in \mathcal{D}} \lvert \lvert x_i - \overline{x}\rvert \rvert ^2$ is the sum of the squared distances of each point in the dataset, $\mathcal{D}$, from the dataset barycenter, $\overline{x}$.
\subsection{Ray-Turi}
With the quantity $\Delta_{k k'}$ defined as above for the Davies-Bouldin index, the Ray-Turi measure is defined
\begin{equation}
{\rm RT} = \frac{1}{N}\frac{{\rm WSS}}{\min_{k < k'} \Delta_{k k'}^2}.
\end{equation}
\subsection{S\_Dbw}
The variances of feature across the dataset are collected together into a vector of size $d$, $\mathcal{V} = ({\rm Var}(V_1),\cdots,{\rm Var}(V_d))$, where $V_i$ is the vector of values for the $i^{\rm th}$ feature across all $N$ data points.   The {\it average scattering} is then defined,
\begin{equation}
\mathcal{S} = \frac{\sum_{k=1}^K \lvert \lvert \mathcal{V}_k\rvert \rvert}{K \lvert \lvert \mathcal{V} \rvert \rvert},
\end{equation}
where $\mathcal{V}_k$ is the variance vector restricted to points belonging to cluster $C_k$.  Next, define the limit value,
\begin{equation}
\sigma = \frac{1}{K} \sqrt{\sum_{k=1}^K \lvert \lvert \mathcal{V}_k\rvert \rvert}.
\end{equation}
The {\it S\_Dbw} index is based on the notion of density: the quantity $\gamma_{k k'}$ for a given point counts the number of points in clusters $C_k$ and $C_{k'}$ that lie within a distance $\sigma$ of the point.  This quantity is evaluated at the midpoint, $H_{k k'}$, and at the barycenters, of the two clusters, $\overline{x}_k$ and $\overline{x}_{k'}$, and used to define the quotient,
\begin{equation}
R_{k k'} = \frac{\gamma_{k k'}(H_{k k'})}{\max (\gamma_{k k'}(\overline{x}_{k} ), \gamma_{k k'}(\overline{x}_{k'}))}.
\end{equation}
The mean of $R_{k k'}$ over all clusters is then added to $\mathcal{S}$ to form the index,
\begin{equation}
S\_Dbw = \mathcal{S} + \frac{2}{K(K-1)}\sum_{k < k'} R_{k k'}.
\end{equation}
\subsection{Silhouette}
The mean distance of a point $x_i$ to other points within the same cluster is 
\begin{equation}
a(i) = \frac{1}{n_k-1} \sum_{x_i,x_j \in C_k} \lvert \lvert x_i - x_j \rvert \vert.
\end{equation}
The mean distance of a point $x_i$ to points in a different cluster, $C_{k'}$,  is $\frac{1}{n_{k'}}\sum_{x_j \in C_{k'}} \lvert \lvert x_i - x_j \rvert \rvert$.  The minimum of these inter-cluster means is denoted $b(i)$.  Each point can then be given a {\it silhouette},
\begin{equation}
s(i) = \frac{b(i) - a(i)}{\max(a(i),b(i))},
\end{equation}
and an overall score defined as the mean silhouette score of a cluster averaged over all clusters,
\begin{equation}
S = \frac{1}{K}\sum_{k=1}^K \frac{1}{n_k}\sum_{x_i \in C_k}s(i).
\end{equation}
\subsection{$\tau$}
The $\tau$ index is a version of Kendall's rank correlation coefficient applied to clustering problems; it is defined in terms of previously introduced quantities as
\begin{equation}
\tau = \frac{s^+ - s^- }{\sqrt{N_B N_W \left(\frac{N_T (N_T-1)}{2}\right)}}.
\end{equation}
\subsection{Trace W}
This index is simply the pooled within-cluster sum of squares,
\begin{equation}
{\rm Tr}\, W = {\rm WSS}.
\end{equation}
\subsection{Wemmert-Gan\c{c}arski}
For a point $x_i$ in cluster $C_k$, define
\begin{equation}
R(i) = \frac{\lvert \lvert x_i - \overline{x}_k\rvert \rvert}{\min_{k' \neq k}\lvert \lvert x_i - \overline{x}_{k'} \rvert \rvert}.
\end{equation}
Then the Wemmert-Gan\c{c}arski index is defined
\begin{equation}
{\rm WG} = \frac{1}{N}\sum_{k=1}^K n_k J_k,
\end{equation}
where $J_k = \max\left(0,1-\frac{1}{n_k}\sum_{x_i \in C_k} R(i)\right)$.
\subsection{Xie-Beni}
The Xie-Beni index is defined
\begin{equation}
{\rm XB} = \frac{1}{N} \frac{{\rm WSS}}{\min_{k < k'} \delta(C_k,C_{k'})^2},
\end{equation}
where $\delta(C_k,C_{k'})$ is defined in Eq. \ref{gdi}.

\bibliography{notes}



\end{document}